\documentclass[runningheads]{llncs}

\usepackage{eccv}
\usepackage{eccvabbrv}
\usepackage{graphicx}
\usepackage{booktabs}
\usepackage[accsupp]{axessibility}
\usepackage{multicol}
\usepackage{multirow}
\usepackage{floatrow}
\usepackage[breaklinks,colorlinks,citecolor=eccvblue]{hyperref}
\usepackage{orcidlink}
\usepackage{colortbl}
\usepackage{wrapfig}

\definecolor{mygray}{rgb}{0.89, 0.93, 0.85}
\definecolor{whitesmoke}{rgb}{0.96, 0.96, 0.96}
\definecolor{timberwolf}{rgb}{0.86, 0.84, 0.82}
\definecolor{darkgreen}{rgb}{0.0, 0.5, 0.0}
\definecolor{lightgray}{rgb}{0.9, 0.9, 0.9}
\definecolor{Mycolor1}{HTML}{BAD8F2}
\definecolor{Mycolor2}{HTML}{E0F0FA}

\newcolumntype{a}{>{\columncolor{mygray}}c}

\begin{document}

\title{CLIP-DPO: Vision-Language Models as a Source of Preference for Fixing Hallucinations in LVLMs} 

\titlerunning{CLIP-DPO}
\authorrunning{Y. Ouali et al.}

\author{Yassine Ouali$^{1}$ \vspace{0.05in} Adrian Bulat$^{1,2}$ \vspace{0.05in} Brais Martinez$^{1}$ \vspace{0.05in} Georgios Tzimiropoulos$^{1,3}$ \vspace{0.1em} \\
{\normalsize $^1$Samsung AI Center Cambridge, UK} \quad
{\normalsize $^2$ Technical University of Iași,  Romania} \quad \\
{\normalsize $^3$ Queen Mary University of London, UK}
}
\institute{}\vspace{-0.1in}

\maketitle

\begin{abstract}
  Despite recent successes, LVLMs or Large Vision Language Models are prone to hallucinating details like objects and their properties or relations, limiting their real-world deployment. To address this and improve their robustness, we present \texttt{CLIP-DPO}, a preference optimization method that leverages contrastively pre-trained Vision-Language (VL) embedding models, such as CLIP, for DPO-based optimization of LVLMs. Unlike prior works tackling LVLM hallucinations, our method does not rely on paid-for APIs, and does not require additional training data or the deployment of other external LVLMs. 
  Instead, starting from the initial pool of supervised fine-tuning data, we generate a diverse set of predictions, which are ranked based on their CLIP image-text similarities, and then filtered using a robust rule-based approach to obtain a set of positive and negative pairs for DPO-based training. 
  We applied \texttt{CLIP-DPO} fine-tuning to the MobileVLM-v2 family of models and to LlaVA-1.5, in all cases observing significant improvements in terms of hallucination reduction over baseline models.
  We also observe better performance for zero-shot classification, suggesting improved grounding capabilities, and verify that the original performance on standard LVLM benchmarks is overall preserved.
\end{abstract}

\section{Introduction}
\label{sec:intro}

A Large Vision Language Model (LVLM) is a Large Language Model (LLM) combined with an independently-trained vision encoder, typically taken from VL embedding models like CLIP~\cite{radford2021learning} and often kept frozen. LVLMs have recently become the state-of-the-art for vision language understanding~\cite{liu2024visual,dai2023instructblip,bai2023qwen,sun2023aligning,zhao2023mmicl}. 
Unlike the prior generation of models~\cite{li2022blip,wang2022git}, which were typically tuned for one or a small number of tasks, LVLMs allow free-form dialog in natural language with an image in the loop. This direction largely draws inspiration from the recent success of ChatGPT~\cite{chatgpt} and respectively, ChatGPT-4V~\cite{gpt4v}, which adapts LLMs to follow human instruction and preferences via various forms of reinforcement learning from human feedback (RFLH) and/or supervised fine-tuning (SFT) on high-quality multi-turn instruction (\ie chat) data. In a similar fashion, the current generation of LVLMs is trained as part of a two-step process~\cite{liu2024visual}. First, the CLIP vision encoder is aligned with the LLM by training a few adaptation layers only and, then, by fine-tuning the model using SFT on a multi-modal instructional dataset. Despite their remarkable success, a major limitation hindering the wider adoption of LVLMs is the high rate of hallucinated details (\eg non-existing objects) that these models exhibit in their generated output text.

\vspace{0.05in}\noindent\textbf{Motivation.} Following the footsteps of LLM training, a natural solution to combat hallucinations is given by RFLH. As reinforcement learning-based training is generally expensive, often requiring the training of a policy model, Rafailov et al.~\cite{rafailov2024direct} proposed a simplified framework, coined Direct Preference optimization (DPO), that allows direct training of the model using a binary cross-entropy objective. The data for DPO is typically obtained either from human preferences or using LLMs to construct preference data automatically. Thanks to its efficiency, DPO has been quickly adopted for LVLMs too~\cite{li2023silkie,zhao2023beyond,li2024multi}. However, these approaches require multiple LVLMs for generation, collecting additional data, and using GPT-4/GPT-4V API for ranking. Moreover, the GPT-4 provided scores are discrete (hence, of reduced granularity) and are themselves prone to hallucinations, which can perpetuate and exacerbate the already high rate of factual errors and poor image grounding~\cite{wang2023llm,li2023evaluating}. Finally, GPT-4/V is behind a paywall, and using it for generation/ranking is neither scalable nor cost-efficient.

\vspace{0.05in}\noindent\textbf{Main idea.} To fix hallucinations and address the high data construction and annotation cost exhibited by current methods, we propose a new DPO variant, called \texttt{CLIP-DPO}, that uses a pre-trained CLIP model~\cite{radford2021learning} to rank the LVLM's self-generated captions to construct positive-negative pairs for DPO. Since the CLIP model was trained in a contrastive manner to measure the alignment between image-text pairs, it is naturally suitable for determining the quality of a given output from an LVLM, grounding it to the correct object or attribute. The dataset over which we operate is constructed by running the original pre-trained target LVLM on its own output obtained using prompting, removing both the need for (i) additional external data and (ii) ensembles of external LVLMs. The final data is filtered before training using robust rule-based filtering. 

\vspace{0.05in}\noindent\textbf{Main results.} We applied \texttt{CLIP-DPO} fine-tuning on top of two state-of-the-art models: MobileVLM-v2 (3 models in total) and LlaVA-1.5 7B.
We find that, in both cases, our approach is effective in reducing hallucinations, outperforming all baseline models (\ie the models without \texttt{CLIP-DPO} fine-tuning) by a significant margin. Importantly, \texttt{CLIP-DPO} significantly outperforms our direct competitor HA-DPO~\cite{zhao2023beyond}, outperforms Qwen-VL~\cite{bai2023qwen} trained on significantly larger datasets (\ie 1.4B samples for pre-training and 77M for multitask training; as opposed to \texttt{CLIP-DPO} training on just 0.7M samples).
Finally, our model's enhanced object grounding capabilities are also illustrated for zero-shot image classification, all without degrading the original performance of the base LVLM model.

\section{Related work}

\noindent \textbf{Large Visual Language Models (LVLMs).} Following the unprecedented success of Large Language Models (LLMs)~\cite{brown2020language,zhang2022opt,chatgpt,touvron2023LLaMA,touvron2023LLaMA2}, several works have recently proposed to build multi-modal capabilities on top of them~\cite{dai2023instructblip,liu2024visual,chu2024mobilevlm,chen2023sharegpt4v}. LLaVA~\cite{liu2024visual} and FROMAGe~\cite{koh2023grounding} directly pass the visual tokens produced by a pre-trained CLIP~\cite{radford2021learning} vision encoder to an LLM, either fine-tuning the LLM or adapting it using LoRA~\cite{hu2021lora}. Notably, LLaVA training includes a pre-training stage that aligns the CLIP features with the LLM input tokens using a simple projection layer (\ie keeping the rest of the model frozen) using image captioning data. InstructBLIP~\cite{dai2023instructblip} uses QFormer~\cite{li2023blip} to reduce the number of vision tokens before passing them to the LLM. As the quality and distribution of the training data plays a crucial role~\cite{zhou2024lima}, a series of methods~\cite{ye2023mplug,chen2023shikra,peng2023kosmos,chen2023sharegpt4v} introduced improved data construction pipelines. For example, ShareGPT4v~\cite{chen2023sharegpt4v} uses the API of GPT-4V~\cite{gpt4v} to first label, then train a model, and finally use it to re-annotate a new training set. Another line of work is improving efficiency by reducing the models' size (original LLaVA models have 7B and 13B parameters). LLaVA-Phi~\cite{zhu2024LLaVA}, MobileVLM~\cite{chu2023mobilevlm} and its follow-up, MobileVLM-v2~\cite{chu2024mobilevlm}, replace the LLaMA~\cite{touvron2023LLaMA} and Vicuna~\cite{chiang2023vicuna} LLMs with the smaller 1.4B and 2.7B variants of MobileLLaMA and Phi~\cite{li2023textbooks}. Our work is orthogonal to the aforementioned methods and does not seek to improve the model's architecture. Instead, we propose \texttt{CLIP-DPO}, an improved training approach for LVLMs based on DPO.

\noindent \textbf{Preference optimization.} Instruction tuning can significantly improve LLMs' perceived output quality and usefulness by aligning their responses to a given task domain or human preferences. This is achieved either by direct fine-tuning on expert data~\cite{chung2022scaling,thoppilan2022lamda,mishra2021cross} or via reinforcement learning~\cite{stiennon2020learning,ziegler2019fine,ouyang2022training}. The latter significantly simplifies the data collection process, but still requires complicated training algorithms based on REINFORCE~\cite{williams1992simple} or PPO~\cite{schulman2017proximal}. Recently, a much-simplified approach was proposed, Direct Preference optimization (DPO)~\cite{rafailov2024direct}, which bypasses the need to train a reward model and allows direct training using a cross-entropy loss. Multiple improved versions for LLMs have been proposed in the meantime~\cite{azar2023general,zhao2023slic,xu2023some}. Following this, a recent wave of works on combining DPO with LVLMs have been proposed~\cite{li2023silkie,zhao2023beyond,li2024multi}. Silkie~\cite{li2023silkie} constructs a multi-modal instructional dataset automatically labeled by GPT-4V. Similarly, HA-DPO~\cite{zhao2023beyond}, aiming to reduce the rate of hallucinations, uses a GPT4 model to label and construct positive-negative pairs with and without hallucinations. The work of~\cite{li2024multi} follows a similar path by using a suite of LLMs and LVLMs (\ie Gemini-Vision~\cite{team2023gemini}) to generate and label the data. These methods are then primarily evaluated on the LLaVA benchmark and PoPE~\cite{wang2023evaluation}. In contrast to the aforementioned works, the proposed \texttt{CLIP-DPO} (i) simplifies the pipeline, (ii) removes the need for paid APIs, (iii) removes the need for additional data, and (iv) removes the need for additional external LVLMs. Instead, we make use of a pre-trained CLIP model to rank the generated outputs from a small pool of efficient LVLMs, and show that the proposed \texttt{CLIP-DPO} training provides significant improvements for fixing LVLM hallucinations and, in general, for enhancing the discriminability and robustness of the model as demonstrated by image classification experiments. 

\begin{figure}[t]
    \centering
        \includegraphics[width=\linewidth]{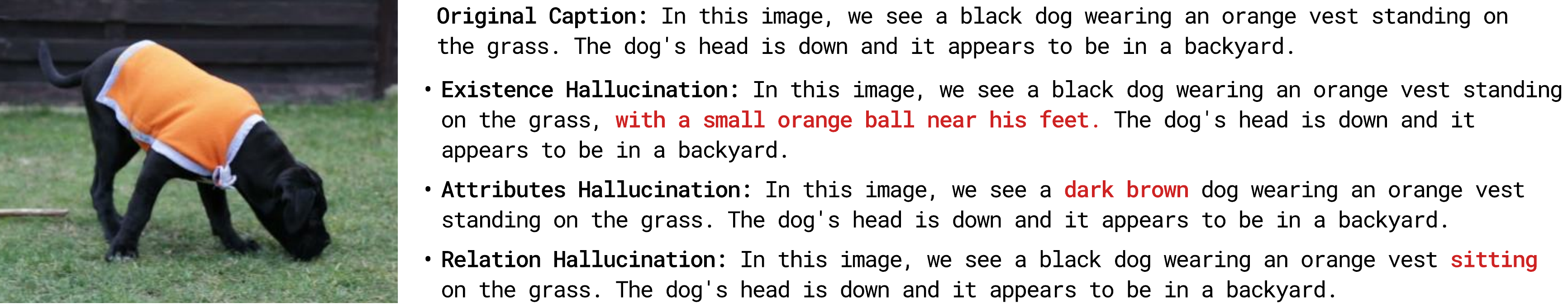}
    \caption{An example of the injected hallucinations. Given an image and its caption, we prompt GPT-4 to generate 3 types of hallucination: existence, attributes, and relation.}
    \label{fig:ablation-hallucination}
\end{figure}

\noindent \textbf{LVLM hallucinations.} Broadly speaking, in the context of LVLMs, we consider hallucinations to be incorrect or misleading generated text, contradicting the visual evidence provided by the input image. This is an undesirable characteristic inherited from the pre-trained LLM used, and further exacerbated by the visual-language alignment process~\cite{hu2023ciem,liu2023mitigating,bai2023qwen,jiang2023hallucination}. Multiple solutions have been recently proposed by the community with varying degrees of success. The works of~\cite{hu2023ciem,you2023ferret} attempt to address the data bias by constructing better-grounded annotated image-text pairs. The works of~\cite{bai2023qwen,chen2023internvl} scale the resolution of the image encoder, as this was observed to reduce the amount of hallucination, but at the cost of a high increase in the computational cost. The works of~\cite{jain2023vcoder,zhai2023halle} improve the vision encoder by adding extra informational paths. The closest work related to ours is that of~\cite{zhao2023beyond}, which, with the help of GPT-4V, constructs negative-positive pairs for DPO fine-tuning. Different from~\cite {zhao2023beyond}, we use a pre-trained CLIP model to perform the ranking, and no additional data or external LVLMs are required. For evaluation purposes, we use AMBER~\cite{wang2023llm}, the most comprehensive benchmark for hallucinations to date, encompassing both a generative and discriminative evaluation component. Importantly, unlike all prior benchmarks~\cite{gunjal2023detecting,liu2023mitigating,jing2023faithscore,sun2023aligning,li2023evaluating}, AMBER is a high-quality dataset fully annotated by humans for both the generative and discriminative tasks.

\section{Preliminaries}

\subsection{The Effectiveness of CLIP in Reducing Hallucinations}
First, we probe the effectiveness of CLIP in accurately ranking correct captions over those with hallucinated content. This assessment will indicate the efficacy of CLIP-style models in reducing hallucinations with DPO-based training. To this end, initially, we select 1K captions from the Detailed Caption~\cite{detailed_caption_dataset_2024} dataset, chosen for its high-quality labels. Then, we instruct GPT-4~\cite{achiam2023gpt} to create three hallucinated captions for each image, corresponding to the following types of hallucination: (i) \textit{existence}: new elements or objects are added to the caption that were not mentioned originally, (ii) \textit{attribute}: the attributes, characteristics, or features of the original caption's elements are altered, and (iii) \textit{relationship}: the spatial, contextual, or interactive relationships of the original caption's elements are altered. See Fig.~\ref{fig:ablation-hallucination} for an example of the injected hallucinations.

Next, we use LLaVA-1.5 7B\cite{liu2023improved} to compute the likelihood of all captions, including the original and the hallucinated captions. 
We then retain only the hallucinated captions for which the model predicts a higher likelihood than the original, yielding 61 for existence, 76 for attribute, and 119 for relationship hallucinations out of the initial 1K captions. Finally, we compute the CLIP

\begin{wrapfigure}[16]{r}{0.48\textwidth}
    \centering
    \vspace{-0.8cm}
    \includegraphics[width=0.8\linewidth]{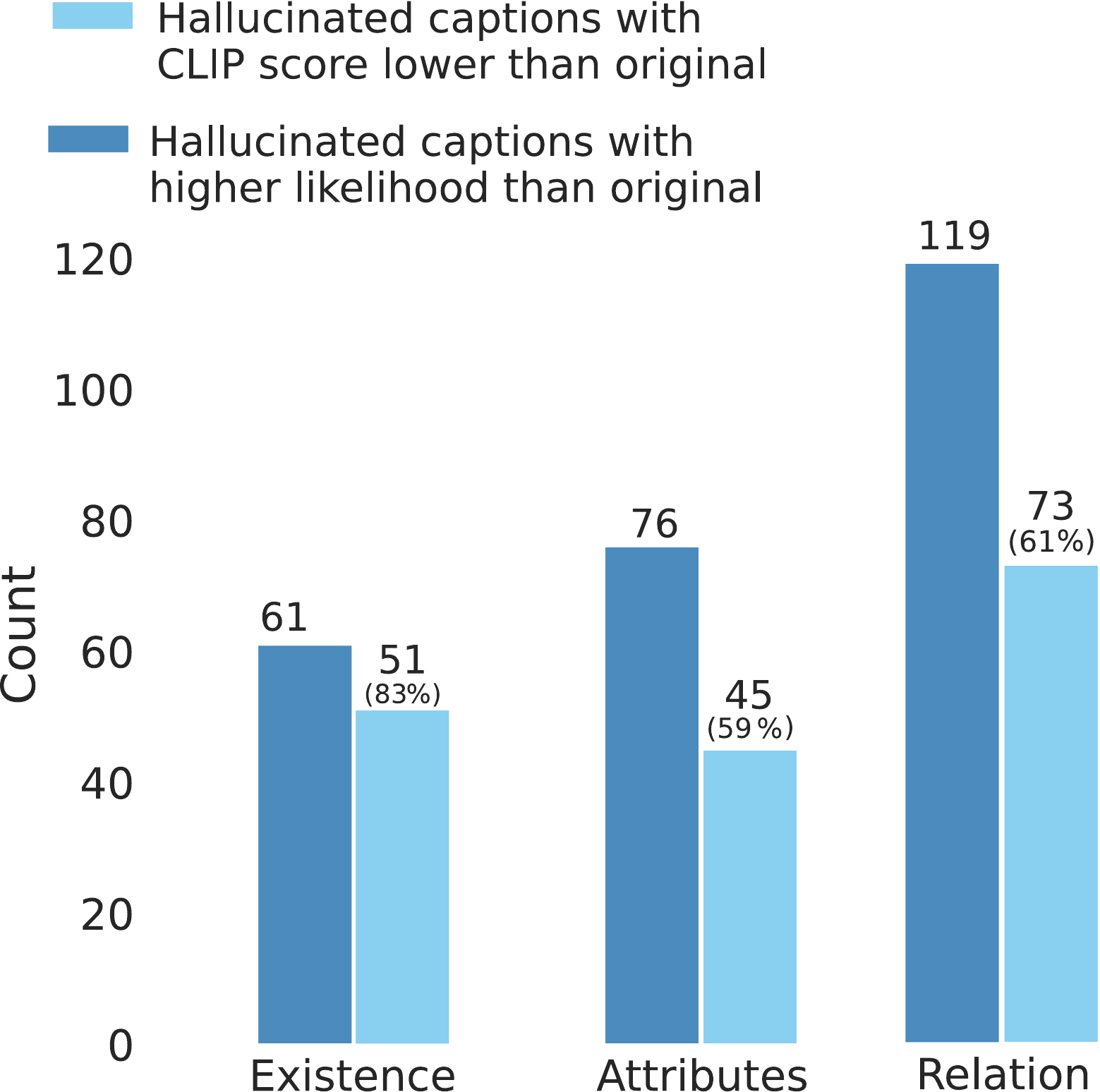}
    \caption{In dark blue, we show the number of hallucinated captions per type that LLaVA-1.5 7B assigns a higher likelihood than the original out of 1K captions. Light blue shows the portion of samples corrected by CLIP.}
    \label{fig:ablation-clip-hallucination}
\end{wrapfigure}
\noindent image-text scores for these likely yet incorrect captions. As illustrated in Fig.\ref{fig:ablation-clip-hallucination}, CLIP manages to accurately rank the original caption over the hallucinated ones in at least 59\% of cases, reaching up to 83\% for the existence type of hallucinations. These findings underscore the potential of using VL embedding models to provide a reliable training signal for hallucination reduction in LVLMs.

\subsection{Direct Preference Optimization}
The standard pipeline for LLM and LVLM alignment consists of three steps: generation, annotation, and
optimization. First, a set of $N$ prompts $\{x_i\}_{i=1}^N$, each one
consisting of an image and a text component in the case of LVLMs, are used to generate a set of response pairs $y^1_i$ and $y^2_i$ obtained from a pool of pre-trained LVLMs for each prompt $x_i$. Then, either human annotators or another set of LLMs (\ie AI annotators)
are used to rank the responses, resulting in a preferred $y^{+}_i$ and a less preferred
response $y^{-}_i$ for each prompt $x_i$, 
and thus a final preference dataset $\mathcal{D}=\{(x_i, y^{+}_i, y^{-}_i)\}_{i=1}^N$.
Finally, Direct Preference Optimization~\cite{rafailov2024direct} (DPO)
can be applied to update the target policy $\pi_\theta$ parameterized by $\theta$ directly
using the preference dataset $\mathcal{D}$. Specifically, the DPO optimization objective is defined as follows:
\begin{equation}
\label{eq:dpo}
    \max _{\pi_\theta} \mathbb{E}_{(x_i, y^{+}_i, y^{-}_i) \sim \mathcal{D}} \log \sigma\left(\beta \log \frac{\pi_{\theta}( y^{+}_i \mid x_i)}{\pi_{\mathrm{ref}}( y^{+}_i \mid x_i)}-\beta 
    \log \frac{\pi_{\theta}(y^{-}_i \mid x_i)}{\pi_{\mathrm{ref}}(y^{-}_i \mid x_i)}\right),
\end{equation}
with $\pi_{\theta}$ as the policy to be learned, $\pi_{\mathrm{ref}}$ as the reference
SFT policy, and $\beta$ as a hyperparameter to control the Kullback-Leibler (KL) divergence between the learned $\pi_{\theta}$ and reference $\pi_{\mathrm{ref}}$ policies~\cite{rafailov2024direct}.
The main benefit of DPO-based optimization is the direct alignment of the LVLM towards the preferences
implicit in the preference data $\mathcal{D}$.

In contrast to previous DPO-based LVLM preference optimization works~\cite{li2023silkie,zhao2023beyond,li2024multi}, next, we will introduce the proposed \texttt{CLIP-DPO}, which (i) simplifies the generation process by limiting the pool of LVLMs used for generation to small and efficient models, 
(ii) removes the need for external LLMs and LVLMs annotators accessed via paid APIs (\eg GPT-3.5, GPT-4 or GPT-4V) 
by using CLIP as the ranking model, and (iii) removes the need for additional data by reusing the same data used during the SFT step.

\begin{figure}[ht!]
    \centering
    \includegraphics[width=1.0\textwidth]{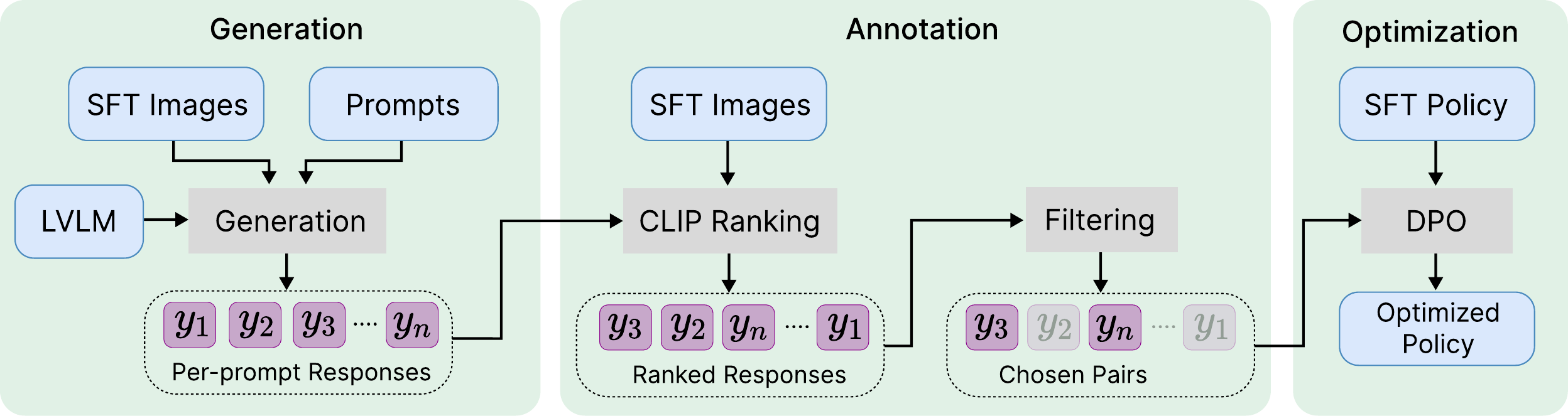}
    \caption{\texttt{CLIP-DPO}. Starting from the initial SFT data pool and a set of prompts, an LVLM generates captions. These captions are first ranked using a CLIP model, then filtered to identify the most suitable positive and negative pairs for DPO-based optimization.} %
    \label{fig:main-idea}
\end{figure} 

\section{Method}

In this section, we introduce our DPO variant for LVLMs preference optimization, named \texttt{CLIP-DPO}. As illustrated in Fig.~\ref{fig:main-idea}, \texttt{CLIP-DPO} follows a three-step process: generation, annotation, and optimization, similar to the standard DPO-based training pipeline. While the final DPO-based optimization step remains unchanged, \texttt{CLIP-DPO} modifies the first two steps. 

Firstly, we streamline the generation step by utilizing the same data pool (\ie images only) as the SFT stage and generate data using either the model itself or a small pool of efficient LVLMs. Secondly, we introduce an annotation step tailored for CLIP-style embedding models, where we rank captions based on their image-text similarities. Then, given that the resulting captions and pairs may vary in quality, we implement a rule-based filtering strategy to remove unsuitable candidates. Finally, with the constructed preference data, the model undergoes DPO-based optimization using Eq.~\ref{eq:dpo}, as shown in Fig.~\ref{fig:main-idea}.

As such, by leveraging VL embedding models for data annotation, we tap into the extensive information captured by these contrastively-pretrained models, which have been exposed to hundreds of millions of unique image-text pairs during training, unlike existing open-sourced LVLMs. Additionally, unlike the discrete GPT-4V ranking used in prior work, CLIP scores are continuous, enabling finer-grained comparisons. This allows us to assess the difficulty of a given pair based on the margin between the score values.

Next, we will elaborate on each module of our pipeline. Sec.~\ref{ssec:data-generation} details the data generation module, and Sec.~\ref{ssec:data-annotation} details the data annotation process.

\subsection{Data Generation}\label{ssec:data-generation}

The first step of our pipeline consists of generating a set of per-image captions that will be later ranked by a pre-trained CLIP model, along with a subsequent filtering approach
to select a set of positive and negative pairs, which are then used for DPO-based training. To this end, we start by selecting the pool of LVLM annotators and the data to be annotated.
For the annotators and to reduce the cost of the generation step, we select MobileVLM-v2~\cite{chu2024mobilevlm} family of models, given their efficiency and performance.
As for the data, and to avoid introducing any new sources that might bias the results of \texttt{CLIP-DPO}, and to further reduce the method's cost, 
we opt for the initial pool of data used during the SFT stage of MobileVLM-v2~\cite{chu2024mobilevlm} models, see Table~\ref{tab:data_splits} for details. Then,
we conduct our data generation pipeline that consists of two steps, the generation of \textit{generic captions} and \textit{per-image questions and answers}.

\hspace{-0.25in}
\begin{minipage}{\textwidth}
    \begin{minipage}[b]{0.48\textwidth}
    \centering
    \scalebox{0.6}{
    \begin{tabular}{l@{\hskip 1.0in}l}
    \toprule
        \textbf{Dataset} & \textbf{Dataset Size} \\
        \midrule
        COCO & 118K \\
        GQA & 72K \\
        LLaVA-1.0 Pretraining & 595K \\
        LLaVA-1.5 Pretraining & 558K \\
        OCR VQA & 80K \\
        SAM & 570K \\
        SBU & 845K \\
        TextVQA & 22K \\
        VG & 86K \\
        Web-Celebrity & 495 \\
        Web-Landmark & 500 \\
        Wikiart & 500 \\
        \midrule
        Total & 2.9M \\
    \bottomrule
    \end{tabular} }
      \captionof{table}{Data pool used for data generation.}
      \label{tab:data_splits}
    \end{minipage}
  \hspace{0.05in}
  \begin{minipage}[b]{0.48\textwidth}
    \centering
    \includegraphics[width=0.8\linewidth]{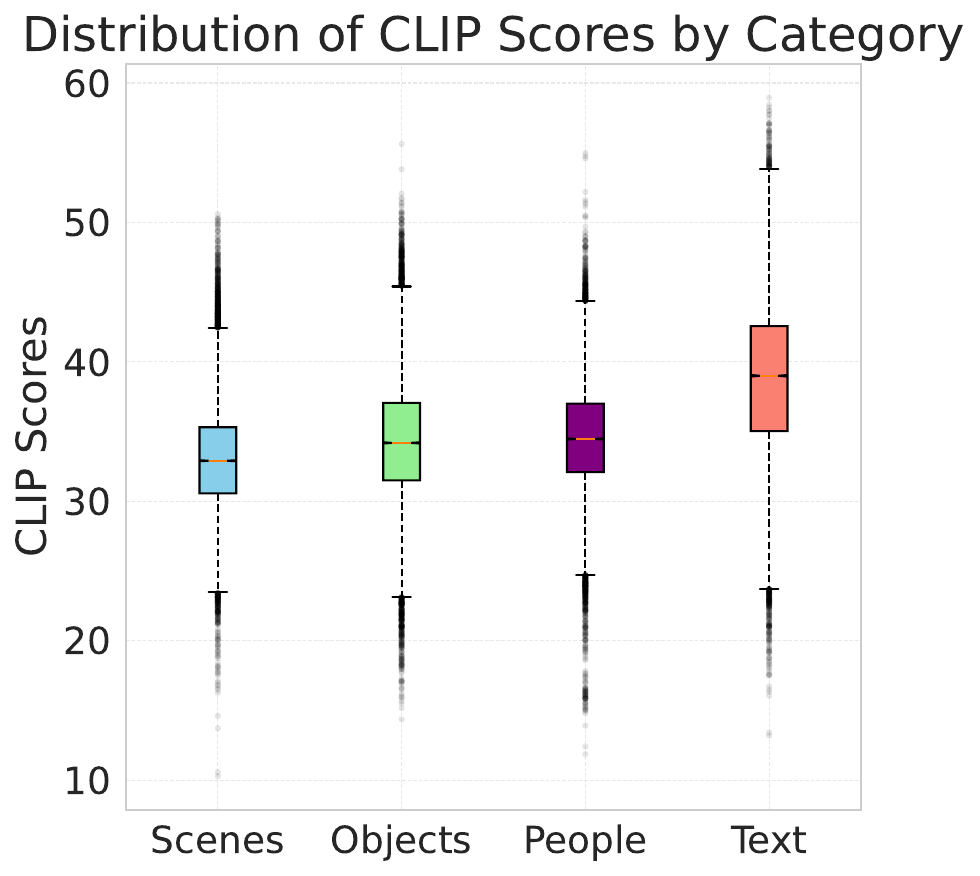}
   \captionof{figure}{The distribution of CLIP image-text scores per category.}
   \label{fig:distribution_of_cats}
  \end{minipage}
\end{minipage}\vspace{-0.1in}

\begin{figure}[ht]
    \centering
        \includegraphics[width=\linewidth]{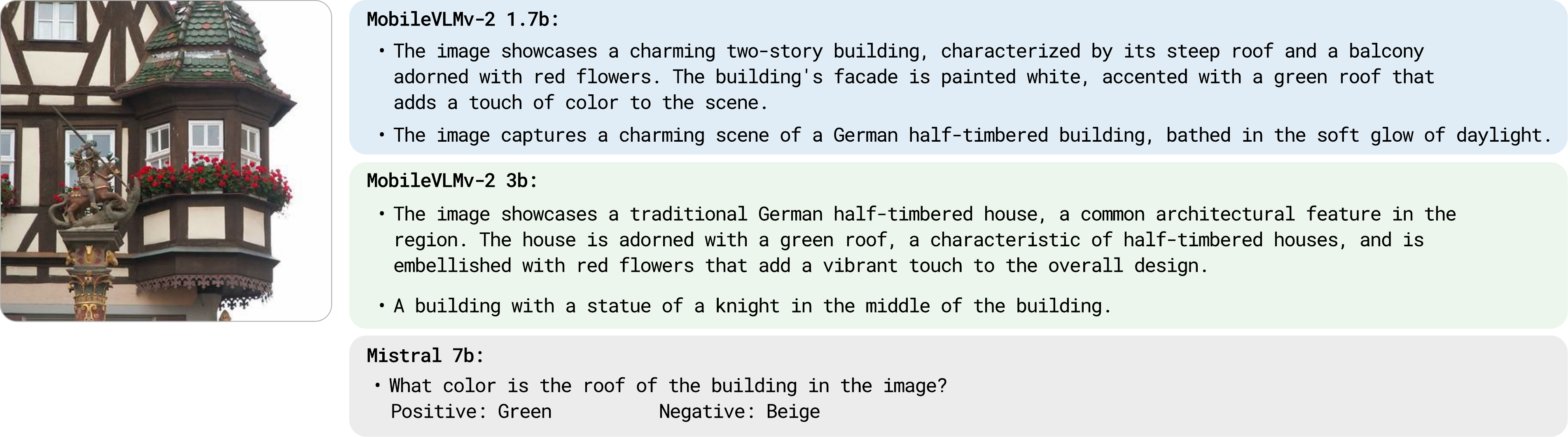}
    \caption{Examples of generated generic captions produced by MobileVLM-v2 1.7B and MobileVLM-v2 3B. For MobileVLM-v2 3B's generated captions, we also show the produced question and positive and negative answers generated using Mistral 7B.}
    \label{fig:data-annotation}
\end{figure}\vspace{-0.25in}

\vspace{0.1in}\noindent\textbf{Generation of Generic Captions.}
We start by generating a set of 5 descriptive captions per image. %
For each of the MobileVLM-v2 models, we prompt with 5 different prompts (\eg "identify the setting and note any characters or objects, focusing on visible details.") to increase the diversity of the generated captions.
While these captions can be used for CLIP ranking and DPO-based training, they are still produced by generic prompts that are not image-specific. Next, we perform the second step of our data generation pipeline to obtain a set of per-image questions and answers.

\vspace{0.1in}\noindent\textbf{Generation of Per-Image Questions and Answers:}
To obtain a set of questions per image, we leverage an LLM, \ie
a variant of Mistral 7B Instruct-v0.2~\cite{jiang2023mistral}\footnote{\url{https://huggingface.co/teknium/OpenHermes-2.5-Mistral 7B}}, and feed it with the generated captions and prompt it to generate 2 questions for each image, together with positive and negative answers. The LLM is asked to generate the positive ones based on the fed captions and to generate plausible but incorrect negatives given the image description/caption.

\vspace{0.1in}\noindent For an example of the generated captions, questions, and answers, see Fig.~\ref{fig:data-annotation}.

\subsection{Data Annotation}\label{ssec:data-annotation}

\begin{figure}[t]
    \centering
        \includegraphics[width=\linewidth]{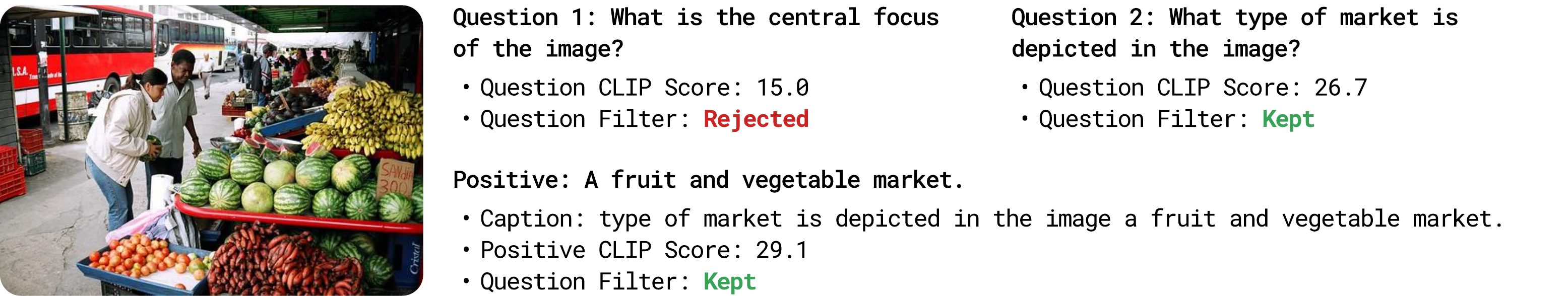}
    \caption{Question Filtering. The sample is rejected if the CLIP image-question score is low for a given sample. If the question is relevant and not generic, it is kept. Next, we evaluate the quality of the answer by first parsing the question into a caption,
    appending the positive answer to it, and computing its CLIP score. If it is a high score, the sample is kept for DPO-based training.}
    \label{fig:question_filtering}
\end{figure}
\begin{figure}[t]
    \centering
        \includegraphics[width=\linewidth]{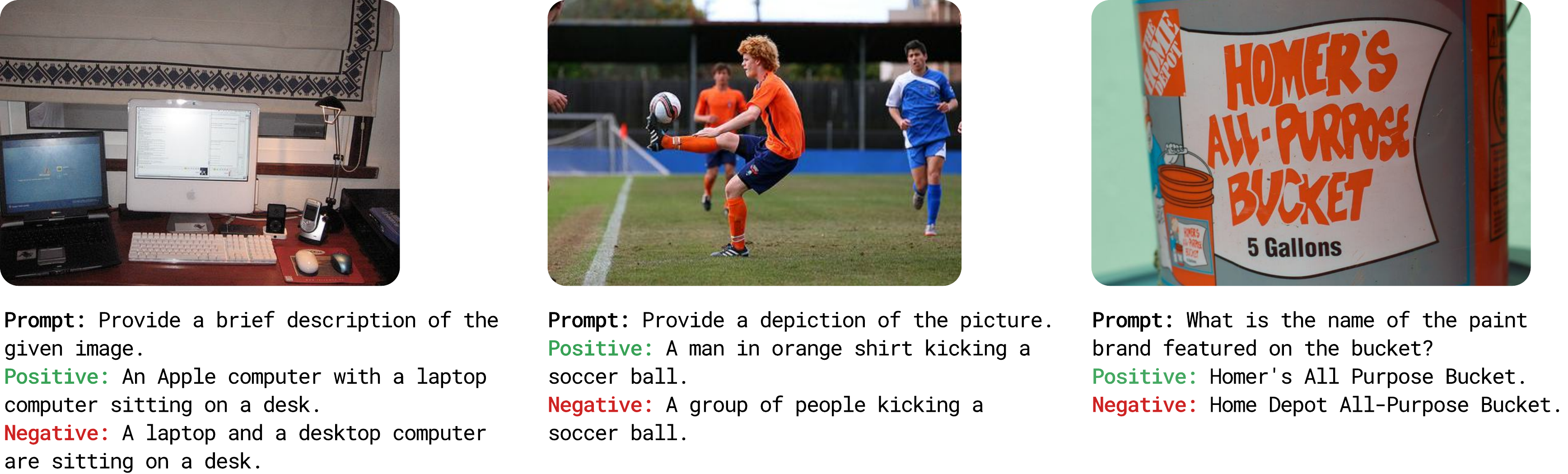}
    \caption{Examples of the final positive and negative pairs obtained post-filtering, to be used for DPO-based training.}
    \label{fig:pairs_examples}
\end{figure}

Starting with the generated captions, the subsequent data annotation step involves ranking them based on their CLIP image-text similarities and filtering them to obtain a final set of high-quality positive-negative pairs for DPO-based training. Our filtering pipeline includes two stages: \textit{global filtering}, where we eliminate low-quality images and captions, and \textit{pair-filtering}, where we select the best positive and negative per-image pairs for DPO-based training.

\vspace{0.1in}\noindent\textbf{CLIP Ranking.} The initial step of data annotation is CLIP ranking. In this step, we use a pre-trained CLIP model to compute the cosine similarities of the captions with their associated images and then sort them from highest to lowest.

\vspace{0.1in}\noindent\textbf{Global Filtering.}
Next, we aim to filter our data and only retain a balanced and higher-quality subset. To achieve this, we start by analyzing the types of images in our data pool (see Tab.~\ref{tab:data_splits}). We define a set of four generic categories: images of text, people, objects, and scenes. Using CLIP and a set of 10 descriptions per category, we create four class prototypes. Each image is then assigned to one of these prototypes based on the highest cosine similarity.
As illustrated in Fig.~\ref{fig:distribution_of_cats}, we observe that CLIP scores for the text category are the highest, while scores for people, objects, and scenes are relatively similar. Consequently, before applying CLIP-based filtering, we down-sample the portion of text images. We then filter out all generic captions below a given CLIP score (\eg < 28.0). Additionally, since CLIP is primarily trained on short text, we remove long captions to ensure more precise CLIP scores.
For questions, we compute their CLIP scores and remove all questions with low CLIP scores (\eg < 25.0), thereby eliminating generic questions (\eg ``what is the main object in the image?'') that are already covered by the generated captions.

\vspace{0.1in}\noindent\textbf{Pair Filtering.}
Finally, starting from the remaining high-quality image captions and image question-answers, the final step will be selecting a set of positive and negative pairs for DPO-based training. For the questions, since we already have a set of positive and negative responses generated by our LLM, we only need to filter out the low-quality pairs. To achieve this, we use simple regex matching rules to extract an image description from the question, append the positive answer to it to create a synthetic caption, and compute its CLIP image-text scores. We then reject examples where the scores are low. See Fig.~\ref{fig:question_filtering} for an example of the questions filtering process.
As for the captions, we consider all possible pairs where the CLIP score difference between two captions is larger than a given threshold (\eg  > 2.0) and where the length of the two captions is similar to avoid introducing false preferences. We then order them based on the CLIP score difference between the positive and negative captions and select the top-ranking pair per image.

\vspace{0.1in}\noindent After the data annotation step, the resulting DPO training data consists of 750K pairs, of which 50K are question-answer pairs and the rest are caption pairs. For examples of the final pairs, see Fig.~\ref{fig:pairs_examples}.

\section{Implementation details}

\noindent \textbf{Network architecture:} In this work, we consider two LVLMs architectures: MobileVLM-v2~\cite{chu2024mobilevlm} and LLaVA-1.5~\cite{liu2024visual}. Both models follow the same overall structure: a pre-trained CLIP vision encoder and a pre-trained LLM. The visual tokens produced by the frozen vision encoder are projected using either a linear layer or a small projection module and passed as input to the LLM. LLaVA opts for a pre-trained Vicuna LLM~\cite{chiang2023vicuna} while MobileVLM-v2 uses MobileLLaMA~\cite{chu2023mobilevlm}, except for their 7B variant, which also uses a Vicuna model. Both use the same ViT-L-14 @ 336px CLIP visual encoder~\cite{radford2021learning}. For efficiency purposes, MobileVLM-v2 uses a projection module that halves the number of visual tokens. We consider the following model variants in our comparisons: MobileVLM-v2 (1.7B, 3B and 7B), and LLaVA-1.5 7B. For data annotation, we use ViT-H/14 DFN~\cite{fang2023data} as our pre-trained CLIP model.

\vspace{0.1in}\noindent \textbf{Training details:} For all of our experiments, unless specified otherwise, we start from the pre-trained MobileVLM-v2 and LLaVA-1.5 models. The models are then fine-tuned for 1 epoch using DPO-based optimization on the constructed \texttt{CLIP-DPO} dataset. We use the following hyperparameters for fine-tuning the models: AdamW optimizer~\cite{loshchilov2017decoupled} with a batch size of 256, a learning rate of $5e-7$, decreased to 0 using a cosine scheduler, a warm-up of 0.01, and a weight decay set to 0. Both during training and testing, the input images are cropped and resized to $336 \times 336$px. The training was performed on 8 A100 GPUs using Pytorch~\cite{paszke2017automatic}. For the larger models, \eg LLaVA-1.5 7B, to fit them in memory, we use the Zero-3 strategy~\cite{rajbhandari2021zero,rasley2020deepspeed} and decrease the batch size to 64.

\section{Results}

We first evaluate the effectiveness of \texttt{CLIP-DPO} in reducing hallucinations on the recently introduced AMBER~\cite{wang2023llm} benchmark in Sec.~\ref{ssec:results-hallucinations}. Moreover, we demonstrate its enhanced grounding capabilities by reporting zero-shot image classification in Sec.~\ref{ssec:results-image-classification-captioning}. Finally, in Sec.~\ref{ssec:results-additional}, we show that the proposed \texttt{CLIP-DPO} training does not compromise LVLM performance by reporting results on the LLaVA benchmark~\cite{zhu2024LLaVA}.

\vspace{-0.3cm}
\subsection{Evaluation on hallucinations}
\label{ssec:results-hallucinations}
We start by evaluating \texttt{CLIP-DPO} in terms of LVLM hallucination reduction, which is our main objective. We use AMBER~\cite{wang2023llm}, a comprehensive, high-quality, and LLM-free multidimensional benchmark for LVLM hallucination evaluation, which can be used to evaluate both generative and discriminative tasks.
As the results from Tab.~\ref{tab:zero_shot_hallucinations} show, MobileVLM-v2 trained with \texttt{CLIP-DPO} improves upon MobileVLM-v2 baselines across all model sizes, and sometimes quite significantly, especially for the 1.7B and 7B models. It can also be seen that \texttt{CLIP-DPO} significantly improves when applied on top of LLaVA-1.5 7B. Importantly, \texttt{CLIP-DPO} significantly outperforms HA-DPO~\cite{zhao2023beyond}, the main competing approach, improving the AMBER score 3.2 vs 7.8 when using LLaVA-1.5. 
Finally, our LlaVA-1.5 7B+\texttt{CLIP-DPO} even outperforms Qwen-VL~\cite{bai2023qwen}, which is trained on significantly larger datasets (1.4B image-text pairs for pre-training and 77M for multitask training while \texttt{CLIP-DPO} is fine-tuned on just 0.7M samples), and even matches the performance of GPT-4V without using any GPT-4V model outputs during training.  
Overall, these results on a high-quality state-of-the-art benchmark such as AMBER clearly demonstrate the effectiveness of our \texttt{CLIP-DPO} approach for reducing hallucinations.

\begin{table*}[t]
    \centering
    \caption{{Hallucination evaluation results} on AMBER for both generative (leftmost set of columns) and discriminative tasks (rightmost set of columns). Our approach offers consistent and large performance improvements across 4 different LVLM models, and also beats HA-DPO by a large margin.}
    \label{tab:zero_shot_hallucinations}
    \resizebox{0.9\textwidth}{!}{%
    \setlength{\tabcolsep}{8pt}
    \begin{tabular}{l| c c c c | c c c c | c}
    \toprule
    \multicolumn{1}{l}{\multirow{2}{*}{\textbf{Model}}}&\multicolumn{4}{c}{\textsc{Generative Task}}&\multicolumn{4}{c}{\textsc{Discriminative Task}}&\multicolumn{1}{c}{\multirow{2}{*}{AMBER}}\\
    \cmidrule(lr){2-5}
    \cmidrule(lr){6-9}
    \multicolumn{1}{c}{}&\multicolumn{1}{c}{CHAIR$_{\downarrow}$}&Cover$_{\uparrow}$&Hal$_{\downarrow}$&\multicolumn{1}{c}{Cog$_{\downarrow}$}&\multicolumn{1}{c}{Acc.}&P.&R.&\multicolumn{1}{c}{F1}&\multicolumn{1}{c}{}\\
    \midrule
    mPLUG-Owl&21.6&50.1&76.1&11.5&40.1&92.8&10.5&18.9&48.7\\
    LLaVA&11.5&51.0&48.8&5.5&42.7&74.1&21.0&32.7&60.6\\
    MiniGPT-4&13.6&63.0&65.3&11.3&63.6&90.5&50.4&64.7&75.6\\
    
    CogVLM&5.6&57.2&\textbf{23.6}&\textbf{1.3}&69.0&88.9&60.9&72.3&83.4\\
    mPLUG-Owl2&10.6&52.0&39.9&4.5&75.6 & \textbf{95.0} &66.9&78.5&84.0\\
    InstructBLIP&8.8&52.2&38.2&4.4&76.5&84.5&79.0&81.7&86.5\\
    Qwen-VL&5.5&49.4&\textbf{23.6}&1.9&81.2&90.8&79.7&84.9&89.7\\
    GPT-4V&\textbf{4.6}&\textbf{67.1}&30.7&2.6&\textbf{83.4}&84.9&\textbf{90.1}&\textbf{87.4}&\textbf{91.4}\\
    
    \midrule
    LLaVA-1.5 7B &7.8&\textbf{51.0}&36.4&4.2&72.0&\textbf{93.2}&62.4&74.7&83.5\\
    \ + HA-DPO&7.2&33.6&19.7&2.6&68.3&68.1&\textbf{98.4}&80.5&86.7\\
    \rowcolor{Mycolor2} \ + CLIP-DPO & \textbf{3.7} & 47.8 &\textbf{16.6}&\textbf{1.3}&\textbf{77.8}&84.4&81.5&\textbf{82.9}&\textbf{89.6}\\
    \midrule
        MobileVLM-v2 1.7B &\textbf{3.8} &	39.6 & \textbf{8.9} & \textbf{0.5} & 65.4 & \textbf{92.6}  & 51.9  & 66.5 & 81.4 \\
         \rowcolor{Mycolor2} \ + CLIP-DPO & 4.2 & 38.9 & 10.8 & 0.5 & 71.2 & 88.7 & 64.8 & 74.9 & \textbf{85.3} \\
    \midrule
        MobileVLM-v2 3B & 4.8 &	38.7 & \textbf{11.1} & 0.7 & 73.5 & \textbf{92.1} & 65.7 & 76.7 & 86.0 \\
         \rowcolor{Mycolor2} \ + CLIP-DPO & \textbf{4.7} & \textbf{41.5}  & 13.1 & \textbf{0.5} & \textbf{76.7} & 91.0 & \textbf{71.8} & \textbf{80.3} &	\textbf{87.8} \\
        \midrule
        MobileVLM-v2 7B & 4.4 &	\textbf{38.9} & 10.4 & 0.6 & 71.9 & \textbf{95.0}  & 60.8  & 74.1 & 84.9 \\
         \rowcolor{Mycolor2} \ + CLIP-DPO & \textbf{4.0} & 38.0 & \textbf{10.1} & \textbf{0.5} & \textbf{77.3} & 93.4 & \textbf{70.8} & \textbf{80.5} & \textbf{88.3} \\
    \bottomrule
    \end{tabular}}
\end{table*}

\begin{table}[t]
    \caption{{Zero-shot} image recognition results in terms of Top-1 accuracy (\%). While HA-DPO shows very similar performance to the base model, our \texttt{CLIP-DPO} improves the base model by a very large margin in all cases.}
    \label{tab:zero_shot_image_recognition}
    \centering
    \resizebox{\textwidth}{!}{%
    \begin{tabular}{l|ccccccccc|c}
    \toprule
        Method & StanfordCars & OxfordPets 	& OxfordFlowers & 	Imagenet & 	Food-101 & 	Eurosat & Caltech-101  & UCF-101 & 	SUN397 & Avg  \\
         \midrule
         LLaVA-1.5& 23.2 & 	34.0 & 	7.3 & 	37.9 & 	45.3 & 	49.5 & 	82.7 & 	50.5 & 	43.1 & 	40.0\\
         \ + HA-DPO & 23.2 &	33.4 & 7.1 & 	36.9 &	45.1 & 	49.4 & 	82.7 & 	50.3 & 	42.7 & 	39.7\\
         \rowcolor{Mycolor2} \ + CLIP-DPO & \textbf{30.1} & \textbf{44.8} & \textbf{16.3} & \textbf{42.2} & \textbf{53.9} & \textbf{52.1} & \textbf{84.8} & \textbf{53.1} & \textbf{48.8} & \textbf{47.4} \\
        \midrule
         MobileVLM-v2 1.7B& 17.1 & 	15.2 &	14.5 & 	32.6  & 31.2 &	49.3 & 	77.9 & 	45.5 & 	39.3 & 	33.9\\
         \rowcolor{Mycolor2} \ + CLIP-DPO & \textbf{19.2} & \textbf{32.6} & \textbf{23.8} & \textbf{41.3} & \textbf{47.4} & \textbf{48.9} & \textbf{80.8} & \textbf{50.2} & \textbf{49.3} & \textbf{43.7}  \\
        \midrule
         MobileVLM-v2 3B& 14.8 & 23.5 & 9.1 & 35.7 & 38.8 & 53.2 & 81.8 & 	48.6 & 42.0 & 36.7\\
         \rowcolor{Mycolor2} \ + CLIP-DPO & \textbf{28.6} & \textbf{40.9} & \textbf{19.5} & \textbf{44.3} & \textbf{52.0} & \textbf{56.4} & \textbf{85.5} & \textbf{52.3} & \textbf{50.1} & \textbf{47.7} \\
        \midrule
         MobileVLM-v2 7B& 27.1 & 32.5 & 	11.5 & 	37.7 & 	45.0 & 	43.1 & 	81.9 & 49.0 & 	46.6 & 41.6 \\
         \rowcolor{Mycolor2} \ + CLIP-DPO &  \textbf{32.5} &  	\textbf{51.3} & \textbf{35.8} & \textbf{50.1} & \textbf{62.6} & \textbf{59.3} & \textbf{88.3} &	\textbf{56.2} & \textbf{54.8} & \textbf{54.5} \\
    \bottomrule
    \end{tabular}
    }
\end{table}

\subsection{Zero-shot image classification }\label{ssec:results-image-classification-captioning}
A primary reason for the hallucinatory behavior of LVLMs is the weak alignment between their visual features and the input LLM tokens. A direct way to evaluate this is through simple zero-shot image classification, which is the go-to benchmark for contrastively trained VL models like CLIP~\cite{radford2021learning}. The typical setup follows a closed-set classification problem, where the names of all possible classes are known a priori and are encoded into class prototypes using CLIP. A given image is then assigned to the class with the highest CLIP image-text scores. Herein, we follow the same protocol with the notable difference that, as our goal is LVLM evaluation, we first prompt the model to generate a free-form image caption describing the main object in the image, then encode the caption using the CLIP text encoder and assign
the image to the class with the highest text-text (\ie caption-class) CLIP score.
Here, we opted for a different family of VL embedding model, \ie SigLIP~\cite{zhai2023sigmoid}, to avoid evaluating using CLIP models similar to those used during the data annotation step.
Following~\cite{radford2021learning,zhou2022conditional}, we evaluate our approach on a suite of 9 diverse datasets: UCF-101~\cite{soomro2012ucf101}, SUN397~\cite{xiao2010sun}, Stanford Cars~\cite{krause20133d}, Oxford Pets~\cite{parkhi2012cats}, Oxford flowers~\cite{nilsback2008automated}, ImageNet~\cite{deng2009imagenet}, Food 101~\cite{bossard2014food}, Eurosat~\cite{helber2019eurosat} and Caltech-101~\cite{fei2004learning}.
As the results from Tab.~\ref{tab:zero_shot_image_recognition} show, all LVLMs fine-tuned with \texttt{CLIP-DPO} significantly outperform their corresponding baselines, showcasing the increased discriminative properties. Note again that these improvements are obtained without affecting the model's performance on other tasks and datasets.

\subsection{Additional evaluations}\label{ssec:results-additional}

Herein, we evaluate the impact of \texttt{CLIP-DPO} training on other vision language tasks and, more specifically, on the popular LLaVA-Bench (GQA~\cite{hudson2019gqa}, ScienceQA~\cite{lu2022learn}, TextVQA~\cite{singh2019towards}, MME~\cite{fu2023mme}, MMBench~\cite{liu2023mmbench}). As Tab.~\ref{tab:zero-shot-LLaVA-bench} shows, overall, \texttt{CLIP-DPO} training does not compromise performance. We further note that accuracy improvements on LLaVA-Bench are heavily tied to the addition of extra training data~\cite{chen2023sharegpt4v} or architectural changes~\cite{liu2023improved,lin2024moe,chu2023mobilevlm,chu2024mobilevlm}. As we are not using any additional data \& models nor making architectural changes, it is not surprising that the performance after \texttt{CLIP-DPO} training remains largely in line with that of the original baseline model.

\begin{table*}[t!]
\centering
\small 
\resizebox{0.95\textwidth}{!}{%
\setlength{\tabcolsep}{3pt}
\begin{tabular}{*{1}{l}*{2}{l}|*{7}{c}}
\toprule
Method & LLM & Res. & GQA & SQA$^\text{I}$ & VQA$^\text{T}$ & POPE & MME$^\text{P}$ & MMB$^\text{dev}$ & Avg.\\
\midrule
BLIP-2~\cite{li2023blip} & Vicuna-13B & 224  & 41.0 & 61.0 & 42.5 & 85.3 & 1293.8 & -- & -  \\
InstructBLIP~\cite{dai2023instructblip}& Vicuna-13B & 224 & 49.5  & 63.1 & 50.7 & 78.9 & 1212.8 & -- & -  \\
Shikra~\cite{chen2023shikra}& Vicuna-13B & 224  & -- & -- & -- & -- & -- & 58.8 & --  \\

Openflamingo~\cite{awadalla2023openflamingo} & MPT 7B & 336 & -- & -- & 33.6 & -- & -- & 4.6 & -- \\
Qwen-VL~\cite{bai2023qwen} & Qwen 7B & 448& 59.3 & 67.1 & \textbf{63.8} & -- & 1487.6 & 38.2 & - \\

mPLUG-Owl~\cite{ye2023mplug} & LLaMA 7B & 224  & -- & -- & -- & -- & 967.3 & 49.4 & -- \\
MiniGPT-v2~\cite{chen2023minigpt} & LLaMA 7B & 448 & 60.3 & -- & -- & -- & -- & 12.2 & --\\
MiniGPT-4~\cite{zhu2023minigpt4} & Vicuna 7B & 224  & 32.2 & -- & -- & -- & 581.7 & 23.0 & -- \\
InstructBLIP~\cite{dai2023instructblip} & Vicuna 7B & 224  & 49.2  & 60.5 & 50.1 & -- & -- & 36.0 & -- \\
ShareGPT4V~\cite{chen2023sharegpt4v} & Vicuna 7B & 336  & \textbf{63.3} & 68.4 & 60.4 & \textbf{85.7} & \textbf{1567.4} & \textbf{68.8} & \textbf{70.8} \\

MoE-LLaVA-1.6B$\times$4~\cite{lin2024moe} & StableLM-1.6B& 336 & 60.4& 62.6 & 47.8 &84.3 & 1300.8$^\text{*}$ & 59.4 & 63.3 \\
MoE-LLaVA-2.7B$\times$4~\cite{lin2024moe} & Phi-2.7B& 336 &  61.1& \textbf{68.7} & 50.2 &85.0 & 1396.4$^\text{*}$ & 65.5 & 66.7 \\
\midrule
LLaVA-1.5~\cite{liu2023improved} & Vicuna 7B & 336  & \textbf{62.0} & 66.8 & 58.2 & \textbf{85.9} & \textbf{1510.7} & 64.3 & 68.8 \\
\ + HA-DPO & Vicuna 7B & 336  & 61.9 & \textbf{69.2} & \textbf{58.3} & 84.3 & 1505.6 & \textbf{64.9} & \textbf{69.0} \\
\rowcolor{Mycolor2} \ + CLIP-DPO & Vicuna 7B & 336 & 59.3 & 67.6 & 56.4 & 85.8 & 1468.7 & \textbf{64.9} & 67.9 \\
\midrule
MobileVLM 1.7B~\cite{chu2023mobilevlm} & MobileLLaMA 1.4B & 336& 56.1 & 57.3 & 41.5 & \textbf{84.5} & 1196.2 & 53.2 & 58.7 \\
MobileVLM-v2  1.7B~\cite{chu2024mobilevlm}&  MobileLLaMA 1.4B & 336 &58.3&66.7&\textbf{52.1}&84.3&\textbf{1302.8}&\textbf{57.7} & \textbf{64.2} \\
\rowcolor{Mycolor2} \ + CLIP-DPO &  MobileLLaMA 1.4B & 336 & \textbf{58.6} & \textbf{68.4} & 47.8 & 84.3 & 1331.1 & 57.6 & 63.8 \\
\midrule
MobileVLM- 3B~\cite{chu2023mobilevlm} & MobileLLaMA 2.7B & 336  & 59.0 & 61.2 & 47.5 & 84.9 & 1288.9 & 59.6 & 62.8 \\
 MobileVLM-v2  3B~\cite{chu2024mobilevlm} &  MobileLLaMA 2.7B  &336 &\textbf{61.1}&70.0&\textbf{57.5}&84.7&\textbf{1440.5}&63.2& 68.1\\
\rowcolor{Mycolor2} \ + CLIP-DPO &  MobileLLaMA 2.7B  &336 &60.9&\textbf{72.3}&57.3&\textbf{85.1}&1425.7&\textbf{63.8}& \textbf{68.4}\\
\midrule
 MobileVLM-v2\cite{chu2024mobilevlm}  7B& Vicuna 7B & 336 &\textbf{62.6}&74.8&\textbf{62.3}&\textbf{85.3}&\textbf{1560.7}&69.2&72.1\\
\rowcolor{Mycolor2} \ + CLIP-DPO  7B& Vicuna 7B & 336 &62.5&\textbf{77.3}&62.2&84.7&1539.1&\textbf{70.4}&\textbf{72.3}\\
\bottomrule
\end{tabular}
}
\caption{\textbf{Results} on several vision language tasks from the LlaVA benchmark. Our approach is combined with LVLMs of different sizes, showing consistent performance in all cases.}\label{tab:zero-shot-LLaVA-bench}
\end{table*}

\subsection{Qualitative Results}\label{ssec:results-additional}
\begin{figure}[t!]
    \centering
    \includegraphics[width=0.9\textwidth]{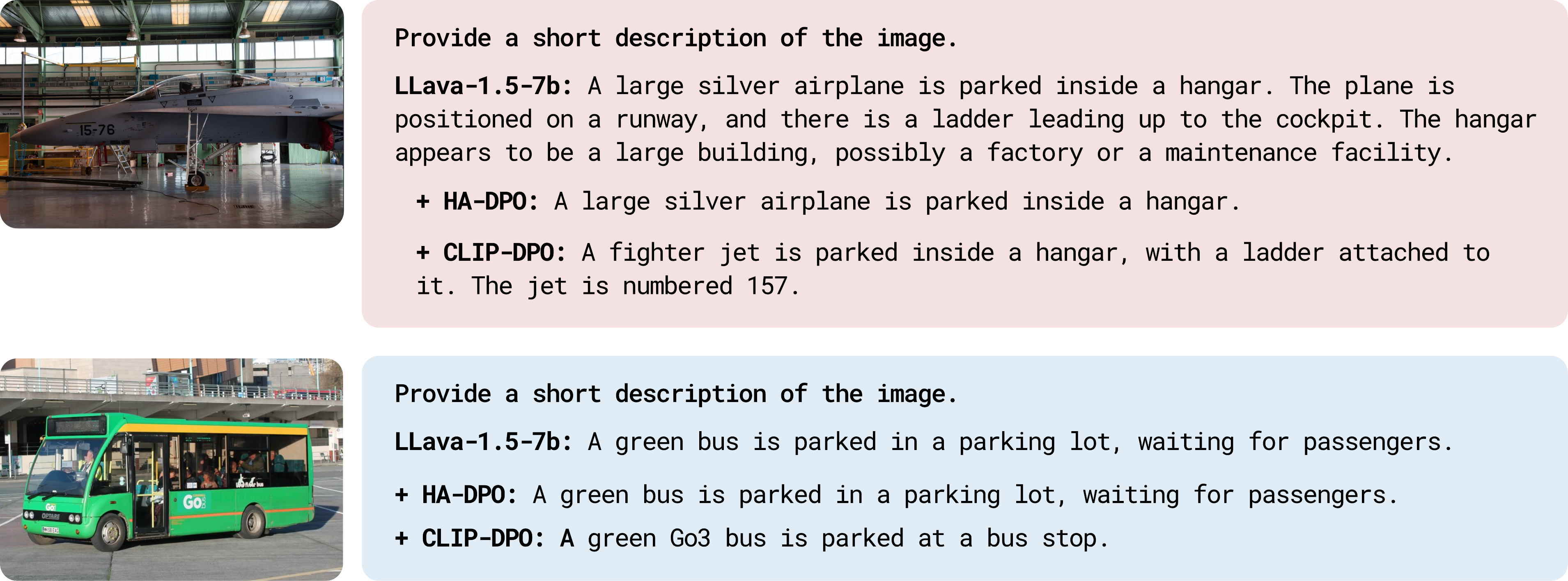}
    \caption{Qualitative examples comparing our method with LLaVA and HA-DPO.}
    \label{fig:qualitative-examples}
\end{figure}

In Figure~\ref{fig:qualitative-examples}, we show some qualitative examples comparing the predictions of
LLaVA-1.5 7B, LLaVA-1.5 7B+HA-DPO, and LLaVA-1.5 7B+\texttt{CLIP-DPO}. Overall, with \texttt{CLIP-DPO}, the model is more grounded in the visual content, less prone to hallucinations, and more precise and fine-grained in its descriptions.

\section{Ablation studies}

Herein, we ablate the effect of the proposed components. In all experiments, we use a MobileVLM-v2 1.7B model.

\subsection{Effect of DPO loss variant}

Our work primarily follows the original DPO formulation proposed in~\cite{rafailov2024direct}. Here we also consider the following recently proposed variations: KTO~\cite{ethayarajh2024kto}, ITO~\cite{azar2023general}, SLIC~\cite{liu2023statistical} and cDPO~\cite{mitchell2024note}.
As Tab.~\ref{tab:ablation-DPO-loss} shows, aggregated, all losses tend to perform similarly, with DPO marginally outperforming the others. 

\begin{table*}[t]
\centering
\caption{\textbf{Effect of DPO loss variant} on \texttt{CLIP-DPO} training. Results are reported in terms of AMBER Score, Average Top-1 (\%) accuracy across all 9 image classification datasets, and the average on LLaVA benchmark. MobileVLM-v2 1.7B was used.}\label{tab:ablation-DPO-loss}
\setlength{\tabcolsep}{3pt}
\scalebox{0.8}{
\small 
\begin{tabular}{l|ccc}
\toprule
Loss variant & AMBER Score & Cls. Avg. & LLaVA bench. \\
\midrule
DPO ($\beta=0.1$)~\cite{rafailov2024direct} & 85.3 & 43.7 & 63.8  \\
DPO ($\beta=0.3$)~\cite{rafailov2024direct} & 83.3 & 42.1 & 64.0  \\
IPO~\cite{azar2023general} &83.6 & 40.5& 64.5 \\
KTO~\cite{ethayarajh2024kto} & 83.5 & 39.5 & 64.7 \\
SLIC~\cite{liu2023statistical} & 83.2 & 42.7 & 64.0 \\
cDPO ($\beta=0.1, LS=0.1$ ~\cite{liu2023statistical}) & 83.1 & 43.2 & 63.9 \\
cDPO ($\beta=0.3, LS=0.1$ ~\cite{liu2023statistical}) & 83.1 & 40.7 & 64.5\\
\bottomrule
\end{tabular}
}
\end{table*}

\begin{table*}[t]
\centering
\caption{\textbf{Effect of CLIP scorer} on \texttt{CLIP-DPO} training. Results are reported in terms of AMBER Score, Average Top-1 (\%) accuracy across all 9 image classification datasets, and the average on LLaVA benchmark. MobileVLM-v2 1.7B was used.}\label{tab:ablation-clip-scorer}
\setlength{\tabcolsep}{3pt}
\scalebox{0.8}{
\small 
\begin{tabular}{l|ccc}
\toprule
Scorer & AMBER Score & Cls. Avg. & LLaVA bench. \\
\midrule
ViT-L/14~\cite{radford2021learning} & 79.8 & 42.3 & 63.0 \\
SigLIP-L/16~\cite{zhai2023sigmoid} & 85.0 & 43.9 & 63.2 \\
ViT-H/14~\cite{schuhmann2022laion} & 84.3 & 44.2 & 63.7 \\
ViT-H/14 DFN~\cite{fang2023data} & 85.3  & 43.7 & 63.8 \\
\bottomrule
\end{tabular}
}
\end{table*}

\subsection{Effect of CLIP scorer}

Herein, we seek to explore alternatives to ViT-H/14 DFN~\cite{fang2023data} CLIP model used in previous experiments, analyzing the impact of the scorer used on the overall performance of the model. We consider a diverse set of alternatives, covering multiple exploratory paths: equally-sized models trained on different data (ViT-H/14 ~\cite{fang2023data} trained on LAION-2B instead of DFN-5B); smaller models (ViT-L/14~\cite{radford2021learning}) and models trained using different pre-training losses (SigLIP-ViT-L/16~\cite{zhai2023sigmoid}). As the results from Tab.~\ref{tab:ablation-clip-scorer} show, our approach is generally robust to the exact scorer used, with the notable exception of the ViT-L/14~\cite{radford2021learning} model. Notice that this difference is primarily manifesting on the AMBER benchmark. Intuitively, this showcases the importance of a powerful scorer for reducing the amount of hallucinations.

\section*{Limitations and broader impact}

As an LVLM-based approach, our method is subject to the same general consideration (\ie potential data bias, susceptibility to hallucinations, etc.). Moreover, as the LVLMs are trained on relatively small datasets compared to LLMs or CLIP, gaps within their knowledge domains are possible. This is especially important as neural networks tend to be overconfident outside their seen input distribution. As with all models from this category, we strongly recommend checking the models and the data carefully before deploying them. Despite these general aspects, our approach is shown to significantly reduce the amount of hallucinated content and improve the model's discriminability, hence resulting in more robust and reliable models.

\section{Conclusions}

In this work, we proposed \texttt{CLIP-DPO}, a simple method that reduces hallucinations in LVLMs based on using a pre-trained CLIP model~\cite{radford2021learning} to rank the LVLM's self-generated captions in order to construct positive-negative pairs for DPO fine-tuning. In contrast to previous works, our proposed \texttt{CLIP-DPO} removes the need of (i) paid-for APIs, (ii) additional data and (iii) additional external LVLMs. Before training, the data is filtered using a newly proposed robust rule-based approach. When applied on top of established LVLMs, it is shown that \texttt{CLIP-DPO} fine-tuning significantly reduces hallucinations, outperforming the baseline models by significant margins and even matching the performance of GPT-4V on the AMBER benchmark. We also observe superior zero-shot object recognition, suggesting improved object grounding capabilities.

\bibliographystyle{splncs04}
\bibliography{main}
\end{document}